\documentclass[runningheads]{llncs}
\usepackage[T1]{fontenc}
\usepackage{graphicx}
\usepackage{float}
\usepackage{ragged2e}
\usepackage{booktabs,tabularx}
\usepackage{multirow}
\usepackage{pdflscape}
\usepackage{makecell}
\usepackage{paralist}
\usepackage{colortbl}
\usepackage{fixltx2e}
\usepackage{color,soul}
\usepackage{xcolor}
\usepackage[rightcaption]{sidecap}
\sidecaptionvpos{table}{c}
\usepackage{hyperref}
\hypersetup{breaklinks=true}
\begin{document}
\title{Placing (Historical) Facts on a Timeline:\\ A Classification cum Coref Resolution Approach}
\titlerunning{Placing (Historical) Facts on a Timeline}
%
\author{\textcolor{black}{Sayantan Adak\orcidID{0000-0001-5307-8811}
\and Altaf Ahmad\orcidID{0000-0002-6211-8237} \and Aditya Basu\orcidID{0000-0003-1004-6507} \and Animesh Mukherjee\orcidID{0000-0003-4534-0044}}}
\authorrunning{S. Adak et al.}
%
\institute{\textcolor{black}{Indian Institute of Technology Kharagpur\\
\email{\{sayantanadak.skni\}@kgpian.iitkgp.ac.in}, 
\email{\{altafahmad3037045\}@iitkgp.ac.in}, 
\email{\{aditya.basu1\}@iitkgp.ac.in},
\email{\{animeshm\}@cse.iitkgp.ac.in}}}

\maketitle              
\begin{abstract}
A timeline provides one of the most effective ways to visualize the important historical facts that occurred over a period of time, presenting the insights that may not be so apparent from reading the equivalent information in textual form. By leveraging generative adversarial learning for important sentence classification and by assimilating knowledge based tags for improving the performance of event coreference resolution we introduce a two staged system for event timeline generation from multiple (historical) text documents. We demonstrate our results on two manually annotated historical text documents. Our results can be extremely helpful for historians, in advancing research in history and in understanding the socio-political landscape of a country as reflected in the writings of famous personas. The dataset and the code are available at  \href{https://github.com/sayantan11995/Event-Timeline-Generation-from-Documents}{https://github.com/sayantan11995/Event-Timeline-Generation-from-Documents}. 
\end{abstract}

\section{Introduction}
Timeline serves as one of the most effective and easiest means to contextualize and visualize a complex situation ranging from grasping spatio-temporal facts in historical studies to critical decision making in businesses. With the stupendous increase of textual resources for many historical contents in several online platforms it has become imperative for the history researchers to understand the chronological orderings of the incessant historical phenomenon. The fact timeline can be an extremely useful aid to highlight the temporal and causal relationships among several facts and the interactions of the characters over time, that results in identifying common themes that arise over the period of interest in a historical document (see Figure~\ref{event_timeline_intro} in Appendix~\ref{eg:timeline}). 

\noindent In this paper we present a full pipeline to build a chronology of facts extracted from historical text. Our contributions are as follows.

\begin{compactitem}
\item \textcolor{black}{We curate a first of its kind dataset from two different historical texts -- the \textit{Collected Works of Mahatma Gandhi} (CWMG) and the \textit{Collected Works of Abraham Lincoln} (CWAL) for our experiments. For each of these datasets we manually annotate sentences that correspond to important facts. Next for each of these annotated sentences we also further annotate the coreferences to the same fact; we call these fact coreferences. Upon acceptance we shall release this data for future research.}
\item \textcolor{black}{We introduce a novel divide-and-conquer based approach to generate fact timeline from timestamped historical texts. In the first step, we classify sentences as containing facts or not using a generative adversarial learning setup. In the subsequent step we compute fact coreferences using both unsupervised and supervised methods. The main novelty here is that inclusion of world knowledge in the form of tag embeddings results in higher performance gains.}
\item \textcolor{black}{We present a rigorous evaluation of both the steps as well as the full system which was absent in previous literature~\cite{bedi-etal-2017-event}. Further we compare our results to the closely related fact timeline summarization tasks by suitably adapting them so that the comparison is fair. 
\item In order to determine the readability and usefulness of the timeline, we conduct an online crowd-sourced survey. 93\% survey participants found it to be effective in summarizing historical timeline of facts.}

\item We also show that our method is generic by evaluating it against a COVID-19 news related dataset which is not a historical text per se. 
\end{compactitem}

\section{Related work}
\textcolor{black}{\noindent\textbf{Important sentence classification \& sentence coreference resolution}: Our proposed approach combines important sentence classification, filtering historically important sentences from a bunch of texts, and sentence coreference resolution, merging factually similar sentences. \cite{zhang2016sensitivity}
used CNN to analyse sensitivity for text
classification. \cite{miyato2017adversarial} and \cite{virtual_adversarial} introduced virtual adversarial training methods for robust text classification from a small number of training data points.\\
Recent works like \cite{choubey-huang-2017-event}, \cite{kenyondean2018resolving} have used neural network based architecture to train their model on benchmark coreference dataset (ECB$+$ \cite{cybulska-vossen-2014-using}). \cite{eeec} attempted to create an end-to-end event coreference resolution system based on the standard KBP dataset\footnote{\url{ https://www.ldc.upenn.edu/collaborations/past-projects/tac-kbp}}.}

\noindent\textbf{Timeline of historical facts}: \cite{bamman-smith-2014-unsupervised} proposed an unsupervised generative model to construct the timeline of biographical life-facts leveraging encyclopaedic resources such as Wikipedia. \cite{recognizing-biographical-section-wikipedia} also uses Wikipedia for timeline construction of historical facts. \cite{bedi-etal-2017-event} attempted to construct a fact timeline from history textbooks considering the sentences having temporal expressions. \cite{palshikar-etal-2019-extraction-message} proposed an automatic approach to capture and visualize temporal ordering of interactions between multiple actors. \cite{gandhipedia} created an AI-enabled web portal based on CWMG dataset. 

\noindent\textbf{Timeline summarization (TLS)}: \textcolor{black}{The timeline summarization task aims to summarize time evolving documents.
\cite{gholipour-ghalandari-ifrim-2020-examining} evaluated existing state-of-the-art methods for news timeline summarization and proposed \textit{datewise} and \textit{clustering} based approaches on the TLS datasets. \cite{born2020} demonstrated the potential of employing several IR methods on TLS tasks based on a large news dataset. \cite{10.1145/3404835.3462954} proposes a new approach by generating date level summaries, and then selecting the most relevant dates for the timeline summarization.}

\noindent\textbf{The present work}: \textcolor{black}{Our paper is closest in spirit to the work done by \cite{bedi-etal-2017-event}. In this paper the authors outlined the challenges related to fact coreference for timeline generation; however, they did not suggest ways to effectively tackle these challenges and, thereby, solve the problem. We close this gap in our paper by proposing an efficient approach to resolve fact coreference. Our work has also close parallels with the fact timeline summarization (TLS) task. Nevertheless, previous TLS researchers mostly worked on the documents containing multiple news articles, which are rich in facts. These works have not focused much on prior fact detection and have not addressed how they can be effectively generalized in historical text documents such as biographies. Our work for the first time shows that fact detection could largely benefit TLS tasks in the context of historical texts.}

\section{Data preparation}

In this section we present the details of the datasets that we prepare for our experiments. We also outline the overall annotation process of these datasets.


\subsection{Datasets}
\label{Dataset}
\noindent\textit{Collected works of Mahatma Gandhi}: 
We leverage the Collected Works of Mahatma Gandhi (CWMG) available at \cite{CWMG}, an assortment of 100 volumes consisting of the books, letters, telegrams written by Mahatma Gandhi and also the compiled writings of the speeches, interviews engaging Gandhi. This data covers many important historical facts within the time period of 1884-1948 in British colonised India. 

\noindent\textit{Collected works of Abraham Lincoln}: 
The second dataset we have use to demonstrate our system is based on the life-long writings of the $16^\textrm{th}$ president of the United States, Abraham Lincoln, formally known as the Collected Works of Abraham Lincoln (CWAL)\footnote{\url{https://quod.lib.umich.edu/l/lincoln/}} comprising a total of 8 volumes.

\noindent\textit{COVID-19 fact dataset}: In addition, to establish the generalizability of the approach, we collect 140 major facts, that happened in India during the COVID-19 pandemic from different sources such as \textit{Wikipedia}\footnote{\url{https://en.wikipedia.org/wiki/COVID-19\_pandemic\_in\_India}}, \textit{Who.int}\footnote{\url{https://www.who.int/india/emergencies/coronavirus-disease-(covid-19)/india-situation-report}} to be placed on a timeline for elegant visualisation using our system.

\subsection{Pre-processing}
\label{data_preprocessing}
\textcolor{black}{
From the 100 volumes of text files from CWMG we first extract all the letters containing the publication dates and recipients name. There were a total of 28531 letters in the entire CWMG. We primarily use the letters for our experiments as we observe that they contain the best temporal account of the facts. 
\if{0}\begin{figure}[H]
     \centering
     \includegraphics[scale=0.25]{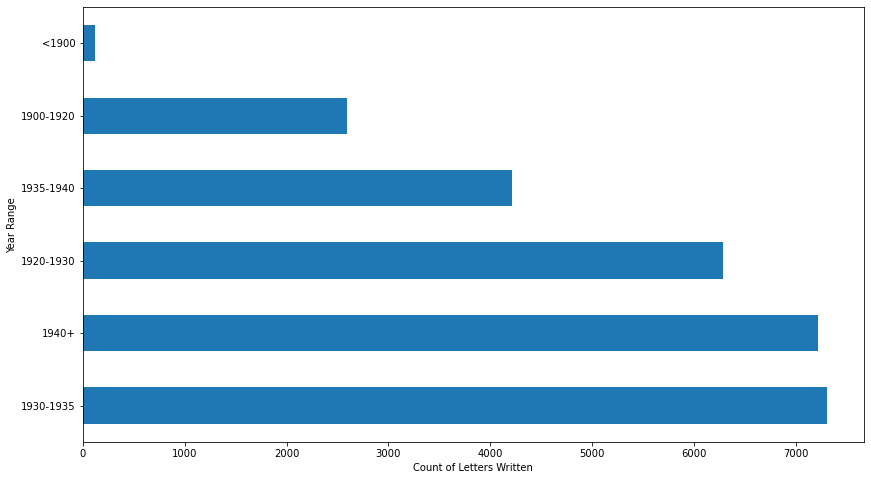}
     \caption{Frequency of letters written by Mahatma Gandhi over the years.}
     \label{letter_frequency}
 \end{figure}\fi
From the overall set of letters, we select the year range 1930--1935 since this range has the largest collection of letters. 
In order to further choose the right data sample, we categorize the letters into \textit{formal} and \textit{informal} types based on the recipients of the letters. A simple heuristic that we follow is -- the letters written to government officials and famous historic personalities can be categorized as formal while those written to the family members can be classified as informal ones. We collect the list of Mahatma Gandhi's family member names from Gandhian experts for identifying the informal letters. We manually notice that the formal letters contain much more useful historic information than the informal ones. We therefore only consider the formal letters for manually annotating the useful sentences. In addition, we only consider the letters which have more than 1000 words in its content. This results in 41 letters with substantial content.}

\begin{table*}
\centering
\scriptsize
\begin{tabular}{  m{.3\textwidth}|m{.4\textwidth}|m{.25\textwidth} } 
 \hline
 \textbf{Doc creation time (Initial reference time)} & \textbf{Important sentences} & \textbf{Updated reference time}\\ 
 \hline
 May 4, 1930 & He was arrested at \hl{12.45 a.m. on May 5.} & May 5, 1930\\ 
 \hline
 May 4, 1930 & In Karachi, Peshawar and Madras the firing would appear to have been unprovoked and unnecessary. & May 4, 1930\\ 
 \hline
\end{tabular}
\caption{\label{sample_sentence}\footnotesize Sample list of sentences from CWMG after the sentence classification. The explicit temporal expression inside the sentence is highlighted.}
\end{table*}

\subsection{Annotation}
\label{Data_annotation}
In this section we outline the data annotation procedure for the two phases. Recall that our method has two important steps -- fact classification and coreference resolution. While the fact classification phase is supervised (Level I annotations), the coreference resolution is done using both unsupervised and supervised techniques. The annotations for the coreference resolution (Level II annotations) are therefore required to (a) train the supervised approach and (b) test the efficacy of both the unsupervised and the supervised approaches. 

\noindent\textit{\textbf{Level I -- Important sentences}}: 
\textcolor{black}{
Finally, out of these filtered letters we manually annotate all the sentences of 18 letters (i.e., 979 sentences in all). The remaining sentences (i.e., 1689 in total) from the rest of the letters were left unlabelled. Both of these labelled and unlabelled sentences were used for training the classifier. The classes in which the sentences were classified were based on their historical importance. In specific, we identify two such important classes -- (a) the \textit{facts} or factful sentences, which typically represent that some important historical phenomena or event \cite{timeML} happened or took place , e.g., `\textit{A vegetable market in Gujarat has been raided because the dealers would not sell vegetables to officials}'\footnote{Such sentences would typically consist of participants and locations.}, 
(b) the \textit{demands}, which represent the demands Mahatma Gandhi had made to the British government through his writings, e.g., `\textit{The terrific pressure of land revenue, which furnishes a large part of the total, must undergo considerable modification in an independent India.}' and (c) others (i.e., not important). As the examples suggest, each individual sentence is annotated as important (i.e., containing a fact/demand) or not. In order to further enrich the dataset we collect gold standard facts related to Mahatma Gandhi from an additional reliable and well maintained resource\footnote{\url{https://www.gandhiheritageportal.org/}}. We obtain 86 additional sentences thus making a total of 1065 (i.e., $979+86$) important sentences (see Table~\ref{tab:class_dist} for the classwise distribution.). }

\begin{table}[ht]
\renewcommand{\arraystretch}{1}
  \centering
  \footnotesize
    \begin{tabular}{|c|c|c|}
    \hline
    \multirow{2}{*}{\textbf{Classes}} & \multicolumn{2}{|c|}{\textbf{Count}} \\
\cline{2-3}    \multicolumn{1}{|c|}{} & \multicolumn{1}{|c|}{\textbf{CWMG}} & \multicolumn{1}{|c|}{\textbf{CWAL}} \\
    \hline
    \textbf{fact} & 716   & 242 \\
    \hline
    \textbf{demand} & 81    & 96 \\
    \hline
    \textbf{other} & 268    & 382 \\
    \hline
    \end{tabular}%
    \caption{\footnotesize Category distribution for the two datasets.}
  \label{tab:class_dist}%
\end{table}%

For the CWAL we simply extract all the sentences from volume 2 and follow similar approaches to annotate important sentences as in the case of CWMG. Without considering any filtering criteria we consider all the 111 articles of volume 2 including his letters and propositions which consist of a total of 1386 sentences. Out of these 720 sentences were manually annotated (see Table~\ref{tab:class_dist}).\\
\noindent\textcolor{black}{\textit{Annotator details and annotation guidelines}: For both the datasets three annotators annotated the sentences. The annotation process was led by one PhD student along with two undergraduate students. The PhD student had substantial experience in historical text analysis and will be referred to as the expert annotator henceforth. The first level of annotation was carried out for each of the sentences and based on the assumption that a full sentence corresponds to a fact/demand. All the annotators annotated the sentences independently. For the training of the two undergraduate annotators, they were provided with the examples of 25 gold standard facts and demands each. The gold standard facts were collected from the reliable resource mentioned in the earlier paragraph and the gold standard demands were collected from the formal letters of Mahatma Gandhi which were first annotated by the expert annotator and verified by a Gandhian scholar (see Table~\ref{sample_annot_1} in  Appendix~\ref{appendix:sample_annot} for example annotations). The inter-annotator agreements, i.e., Cohen's $\kappa$ were 0.66 and 0.58 for the former and the latter datasets respectively. Table \ref{tab:class_dist} shows the category distribution for both the datasets.
The Level I annotation was not carried out for the COVID-19 dataset because, each sentence collected were presented as facts in the mentioned portals and thus we considered all the sentences as important facts.}\\
\noindent\textit{\textbf{Level II -- Coreference resolution}}: \textcolor{black}{The second round of annotation was carried out for evaluating the fact coreference detection task on the same dataset. For this case we only annotate the texts which were marked important during the Level I annotation. In addition, the Level II annotation was also carried out for the COVID-19 fact dataset.}\\
\noindent\textcolor{black}{\textit{Annotator details and annotation guidelines}: The same annotators annotated for the Level II phase. The annotators were provided with sentences, the reference documents (letters) from which the sentences were extracted and the reference time (document publication date). Based on the perception of the annotators, the sentences that potentially referred to the same fact were placed in the same cluster. The coreferences have been placed by the annotators in different clusters based on different factors like the commonness of the mentioned times, entities and the fact name/composition. Consider these two sentences - `\textit{The crowd that demanded restoration of the flag thus illegally seized is reported to have been mercilessly beaten back.}' and `\textit{Bones have been broken, private parts have been squeezed for the purpose of making volunteers give up, to the Government valueless, to the volunteers precious salt}'. Although there is no explicit mention of time in either of the sentences, both of them are from the same document and thus their reference dates would be the same as the publication date of the document. Also both of them refer to similar types of atrocities. So these two sentences should be placed in the same cluster. We first carried out a trial round for the two undergraduate annotators by using 100 randomly chosen important sentences from the Level I phase and the trial annotations were verified by the expert annotator. Finally for the complete Level II annotations, the inter-annotator agreements were 0.74, 0.61, and 0.78 for the CWMG, the CWAL and the COVID-19 dataset respectively using MUC \cite{muc} based F1-score \cite{GHADDAR16.192} (see Table~\ref{sample_annot_2} in  Appendix~\ref{appendix:sample_annot} for example annotations and Appendix \ref{appendix:annot_agree} for other agreement metrics.).} 




\section{Methodology}
\begin{figure*}[ht]
    \centering
    \includegraphics[width=0.95\textwidth]{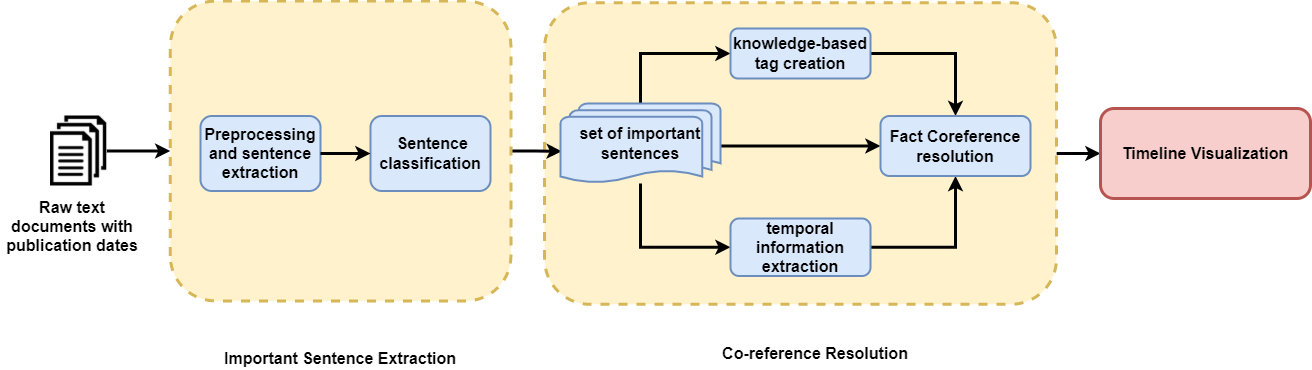}
    \caption{\footnotesize The overall architecture for generating the timeline.}
    \label{architecture}
\end{figure*}

Our method consists of three major components (see Figure~\ref{architecture}): (i) important sentence extraction, (ii) sentence coreference resolution, and (iii) timeline visualization. The arrows represent the direction of data flow. In this section we describe in detail the methods used for each of these components.

\subsection{Important sentence extraction} 

\noindent\textit{\textbf{Baselines}}: As baselines, we use \emph{SVM} \cite{svm} and \emph{Multinomial Na\"ive Bayes} \cite{naive-bayes} on simple bag-of-words feature. For \emph{SVM} we use linear kernel. For the evaluation of the classifiers we use a 70:30 train-test split of the annotated data. 

\noindent\textit{\textbf{Fine-tuned BERT}}: Apart from the above two baselines, we try BERT \cite{devlin2019bert} neural network based framework for the classification. We train the model using the PyTorch~\cite{NEURIPS2019_9015} library, and apply \emph{bert-base-uncased} pre-trained model for text encoding. We use a batch size of 32, sequence length of 80 and learning rate of $2{e-5}$ as the optimal hyper-parameters for training the model. 

\noindent\textit{\textbf{GAN-BERT text classifier}}: In search for further enhancement of the performance based on our limited sets of labelled data, we employ the \emph{GAN-BERT} \cite{croce-etal-2020-gan} deep learning framework for classifying the important sentences. It uses generative adversarial learning to generate augmented labelled data for semi-supervised training of the transformer based BERT model. It improves the performance of BERT when training data is scarce and is therefore highly suited for our case. Here we also feed the unlabeled data sample, as discussed in section \ref{Data_annotation}, to help the network to generalize the representation of input texts for the final classification \cite{croce-etal-2020-gan}.

\subsection{Sentence coreference resolution} 
\label{coreference_resolution_details}
Once the classification was done we end up with 'factful' sentences linked to its corresponding document creation time in the format noted in Table~\ref{sample_sentence}.

\noindent\textit{\textbf{Time within sentences}}: \textcolor{black}{For generating the accurate fact timeline we need to assign a valid date to a particular sentence (i.e fact/demand). For example, in the first sentence in Table~\ref{sample_sentence}, although the document publication time is mentioned to be \texttt{May 4, 1930}, the sentence clearly has embedded in it the exact fact date \texttt{May 5, 1930} apparent from the snippet `\textit{arrested on May 5}'.  Therefore, if the explicit time is present in the sentence we use it directly, else we use the creation/publication date of the document. We extract the explicit mention of time in the text using the \emph{HeidelTime} \cite{strotgen-gertz-2010-heideltime} tool. This tool is capable of identifying embedded mentions of temporal expressions such as \textit{`yesterday', `next day' etc}.}

\if{0}
\begin{figure}
    \centering
    \includegraphics[scale=0.4]{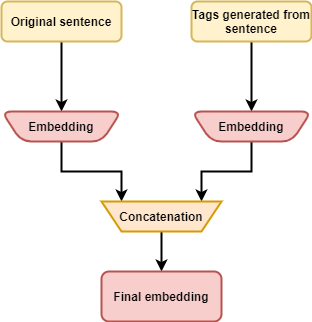}
    \caption{Concatenation of the sentence and the tag embedding through.}
    \label{fig:embedding_concat}
\end{figure}
\fi

\noindent\textit{\textbf{Tag generation from world knowledge}}: An individual sentence does not always contain much information about the fact/demand which it is getting referred to. So we attempt to incorporate world knowledge for each individual sentence. By using each sentence as a query we gather the top five \textit{Google} search results using the \emph{googlsearch} api\footnote{\url{https://github.com/MarioVilas/googlesearch}} and also consider the document from which the sentence was being extracted. Next we analyse the search result using \emph{TextRank}\footnote{\url{https://github.com/DerwenAI/pytextrank}}, \emph{Rake}\footnote{\url{https://pypi.org/project/rake-nltk/}} and \emph{pointwise mutual information}\footnote{\url{https://www.nltk.org/howto/collocations.html}} to generate top keywords present in the search result. Although these methods produce reasonably good results, in many cases we needed to manually filter out certain noisy tags. For each sentence we therefore land up with one or more tags. We retain the top ten tags for every sentence which means that the number of tags for a sentence could vary between one and ten. The details of the tag generation procedure mentioned in Appendix \ref{tag_creation}. We do not use encyclopaedic resources such as Wikipedia to get the search results because the datasets we are using, are only available in a few very specific websites. \textcolor{black}{We fed the list of keyword(s) or tag(s) obtained for a sentence to the pre-trained \emph{sentence-bert} model for obtaining a 768 dimensional embedding representation of the keywords.}

\noindent\textit{\textbf{Unsupervised sentence clustering}}: We employ several unsupervised approaches for sentence coreference resolution. As baselines, we choose two commonly used approaches for coreference resolution -- (a) \emph{Lemma}: It attempts to put the sentence pairs in same coreference chain which share the same head lemma, (b) \emph{Lemma-$\delta$}: In addition to same head lemma as a feature, it also computes the cosine similarity ($\delta$) between the sentence pair based on \textit{tf-idf} features, and only places the sentence pairs in the same coreference chain if $\delta$ exceeds some threshold. Then the sentence clusters were created using agglomerative clustering method. To extract the head lemma of a sentence, we use the \emph{SpaCy} dependency parser.  

Apart from these two common baselines, we vectorize the sentences using \emph{tf-idf} vectorization technique and then apply different clustering techniques such as \emph{Gaussian-Mixture}\footnote{\url{https://scikit-learn.org/stable/modules/mixture.html}} model, \emph{agglomerative clustering} to cluster the sentences corresponding to similar facts. We also use the pre-trained \emph{sentence-bert} \cite{reimers2019sentencebert} model to encode the sentences and apply similar clustering techniques. Finally, we concatenate the sentence embedding with the tag embedding generated from that particular sentence. We again cluster the sentences based on this new representation. This, as we shall later see, significantly improves the performance of the clustering phase. We evaluate the clustering results on the basis of the annotated data which had been obtained in the second phase of data annotation. \textcolor{black}{We used the \emph{elbow} method to find the optimal number of clusters in case of Gaussian-Mixture and used \emph{dendogram} to select the optimal distance threshold for the suitable number of clusters in case of agglomerative clustering. The distance threshold we selected were 0.25, 0.6 and 0.6 for CWMG, CWAL and COVID-19 data respectively.}

\noindent\textit{\textbf{Supervised fact mention-pair model}}: A \textit{fact mention} is a sentence or phrase that defines a fact and one fact may contain multiple \textit{fact mentions} \cite{chen-etal-2009-pairwise}. 
We first create a dataset containing all the possible pairs of \textit{factful} (i.e., fact or demand) sentences from the ground-truth annotations. We set the coreference label to 1 if the sentence pair is contained in the same cluster as per the Level II annotation and 0 otherwise. Here we again use a 70:30 split to generate training and test instances. \textcolor{black}{The overall architecture is inspired from \cite{barhom2019revisiting} (see Appendix \ref{appendix:supervised-model-architecture}).} The inputs to the model are the two sentences (i.e. $S_1$ and $S_2$) and their corresponding \textit{actions} (i.e., $A_1$ and $A_2$), \textit{time} (i.e., $T_1$  and $T_2$) and \textit{tags} (i.e., $K_1$ and $K_2$). We extract \textit{actions} (i.e., $A_i$) for each of the sentences using \textit{SpaCy} dependency parser\footnote{We consider the root verb  as action for a sentence}.


\noindent
\textcolor{black}{\textit{Mention pair construction}: We used \textit{Tensorflow} \cite{tensorflow2015-whitepaper} tokenizer to vectorize each feature (i.e., sentences, actions, time and tags) to convert it into sequence of integers after restricting the tokenizer to use only the top most common 5000 words. For the sentences we limit the sequence length to 64.  For the other features \-- actions, time and tags \-- we limit the sequence length to 10. We always use zero padding for smaller sequences. We next encode the words present in each of these sequences using a pre-trained \textit{GloVe} \cite{pennington2014glove} embedding (100 dimensions). Thus each sentence comes out as a $64*100$ size vector representation while each of the other features come out as a $10*100$ size vector representation. Now each of these vectors are separately passed through a LSTM \cite{10.1162/neco.1997.9.8.1735} layer with default hyperparameters to transform them into 128 size vectors each. Next each of these 128 size vectors are passed through separate dense layers to obtain 32 size vectors. Finally, these 32 size vectors are concatenated using a concatenation layer. The output of the concatenation layer is what we term as a \textit{mention representation}. Two mention representations are concatenated to get a pairwise representation (i.e., an \textit{fact mention pair}) and passed through a feed forward network to return a score denoting the likelihood that two mentions are coreferent (see Figure~\ref{supervised_architecture} in Appendix~\ref{sup_diag}). Based on the predicted pairwise score on the test instances we used a threshold (0.5 in our case) to generate a similarity matrix of the mentions, and then applied agglomerative clustering to partition the similar mentions into the same clusters.}

\subsection{Timeline visualization} 

Once the sentence coreference resolution phase was successfully executed, we generated visualization for the given fact/demand sequence using \emph{vis-timeline}\footnote{\url{https://visjs.github.io/vis-timeline/docs/timeline/}}, a dynamic, browser based visualization library.



\section{Experiments}
\label{sec:exp}
    \subsection{Evaluation metrics}
    We have used separate evaluation metrics for the two phases.
    
 \noindent\textit{Important sentence classification}: In this case we use the standard \emph{accuracy} and \emph{F1-score} values.

\noindent\textit{Sentence coreference resolution}: Here we conduct the evaluation based on the widely used coreference resolution metrics -- (a) \textit{MUC} \cite{muc}, (b) \textit{B\textsuperscript{3}} \cite{bcubed}, (c) \textit{CEAF} \cite{ceaf}, and (d) \textit{BLANC} \cite{blanc}. Due to the inconsistency of each of these evaluation metrics \cite{moosavi-strube-2016-coreference} we shall also report the average outcomes of all the metrics.

    \subsection{Results}
    
    We evaluate the two different phases separately. Ground-truth data was used from each phase for respective evaluations.
    
\noindent\textit{\textbf{Important sentence classification}}: The key results for the two datasets (CWMG and CWAL) are summarised in Table \ref{tab:classification_result}. Our approach based on GAN-BERT by far outperforms the standard baselines. For the CWMG dataset, the macro F1-score shoots from 0.50 (SVM) to 0.69 on the three class classification task. Likewise for the CWAL dataset, the macro F1-score shoots from 0.34 (Na\"ive Bayes) to 0.65.
    
\begin{SCtable}[][h]
\renewcommand{\arraystretch}{1}
  \centering
  \scriptsize
    \begin{tabular}{|c|c|c|c|}
    \hline
    \multicolumn{1}{|c|}{\multirow{2}{*}{Dataset}} & \multirow{2}{*}{Model} & \multicolumn{2}{|c|}{Evaluation Metric} \\
\cline{3-4}          & \multicolumn{1}{c|}{} & \multicolumn{1}{|c|}{Accuracy} & \multicolumn{1}{|c|}{F1} \\
    \hline
    \multicolumn{1}{|c|}{\multirow{4}{*}{\rotatebox[origin=c]{90}{\textbf{CWMG}}}} & MNB   & 0.74  & 0.45 \\
\cline{2-4}          & SVM   & 0.79  & 0.5 \\
\cline{2-4}          & Fine-tuned BERT & 0.8   & 0.57 \\
\cline{2-4}          & GAN-BERT & \cellcolor{green!20}\textbf{0.9}   & \cellcolor{green!20}\textbf{0.69} \\
    \hline
    \multicolumn{1}{|c|}{\multirow{4}{*}{\rotatebox[origin=c]{90}{\textbf{CWAL}}}} & MNB   & 0.6   & 0.3 \\
\cline{2-4}          & SVM   & 0.6   & 0.34 \\
\cline{2-4}          & Fine-tuned BERT & 0.61  & 0.56 \\
\cline{2-4}          & GAN-BERT & \cellcolor{green!20}\textbf{0.7}   & \cellcolor{green!20}\textbf{0.65} \\
    \hline
    \end{tabular}%
\caption{\footnotesize Results (accuracy and macro F1-score) for the important sentence classification using our approaches on the two datasets. MNB: Multinomial Na\"ive Bayes. Best results are marked in boldface and highlighted in green cells.}
  \label{tab:classification_result}%
\end{SCtable}

\begin{table*}[!ht]
\renewcommand{\arraystretch}{.8}
  \centering
  \scriptsize
    \begin{tabular}{|c|c|c|c|c|c|c|c|c|c|}
    \hline
    \multicolumn{1}{|c|}{\multirow{2}{*}{Dataset}} &
      \multirow{2}{*}{System} &
      \multicolumn{1}{|c|}{MUC} &
      \multicolumn{1}{|c|}{B$^{3}$} &
      \multicolumn{1}{|c|}{CEAF\_E} &
      \multicolumn{1}{|c|}{BLANC} &
      \multicolumn{3}{|c|}{Avg (overall)} &
      \multirow{2}{*}{Time taken}
      \\
\cline{3-9}     &
      \multicolumn{1}{c|}{} &
      \multicolumn{1}{|c|}{F1} &
      \multicolumn{1}{|c|}{F1} &
      \multicolumn{1}{|c|}{F1} &
      \multicolumn{1}{|c|}{F1} &
      \multicolumn{1}{|c|}{Recall} &
      \multicolumn{1}{|c|}{Precision} &
      \multicolumn{1}{|c|}{F1} &
      \multicolumn{1}{c|}{}
      \\
    \hline
    \multicolumn{1}{|c|}{\multirow{12}{*}{\rotatebox{90}{\textbf{CWMG}}}} &
      Lemma &
      {0.45} &
      {0.38} &
      {0.20} &
      {0.49} &
      0.39 &
      0.38 &
      0.38 &
      45 sec
      \\
\cline{2-9}   &
      Lemma-$\delta$ &
      0.53 &
      0.41 &
      0.19 &
      0.48 &
      0.48 &
      0.40 &
      0.41 &
      7 min 22 sec
      \\
\cline{2-9}     &
      tf-idf $+$ GM &
      0.53 &
      0.53 &
      0.36 &
      0.60 &
      0.49 &
      0.52 &
      0.50 &
      26 min 14 sec 
      \\
\cline{2-9}     &
      tf-idf $+$ AC &
      0.55 &
      0.50 &
      \cellcolor{blue!20}\underline{0.42} &
      0.57 &
      0.50 &
      0.53 &
      0.51 &
      5 min 13 sec
      \\
\cline{2-9}     &
      s-bert $+$ GM &
      0.61 &
      0.54 &
      0.41 &
      0.60 &
      0.54 &
      0.54 &
      0.54 &
      29 min 34 sec
      \\
\cline{2-9}     &
      s-bert $+$ AC &
      \cellcolor{blue!20}\underline{0.63} &
      \cellcolor{blue!20}\underline{0.57} &
      0.40 &
      \cellcolor{blue!20}\underline{0.61} &
      \cellcolor{blue!20}\underline{0.55} &
      \cellcolor{blue!20}\underline{0.56} &
      \cellcolor{blue!20}\underline{0.55} &
      7 min 42 sec
      \\
\cline{2-10}     &
      \multicolumn{9}{|c|}{$+$ tag embedding}
      \\
\cline{2-10}     &
      tf-idf $+$ GM &
      0.64 &
      0.57 &
      0.45 &
      0.64 &
      0.57 &
      0.60 &
      0.58 &
      28 min 19 sec 
      \\
\cline{2-9}     &
      tf-idf $+$ AC &
      0.62 &
      0.61 &
      0.51 &
      0.66 &
      0.58 &
      0.63 &
      0.60 &
      6 min 57 sec 
      \\
\cline{2-9}     &
      s-bert $+$ GM &
      0.65 &
      0.62 &
      0.48 &
      0.66 &
      0.60 &
      0.60 &
      0.60 &
      30 min 28 sec
      \\
\cline{2-9}     &
      s-bert $+$ AC &
      {0.75} &
      \cellcolor{green!20}\textbf{0.70} &
      {0.52} &
      \cellcolor{green!20}\textbf{0.73} &
      {0.65} &
      \cellcolor{green!20}\textbf{0.71} &
      {0.68} &
      8 min 36 sec
      \\
\cline{2-9}     &
      m-pair model &
      \cellcolor{green!20}\textbf{0.91} &
      {0.59} &
      \cellcolor{green!20}\textbf{0.83} &
      {0.53} &
      \cellcolor{green!20}\textbf{0.83} &
      {0.69} &
      \cellcolor{green!20}\textbf{0.72} &
      2 hr 10 min 32 sec
      \\
    \hline
    \hline
    \multicolumn{1}{|c|}{\multirow{12}{*}{\rotatebox[origin=c]{90}{\textbf{CWAL}}}} &
    Lemma &
      0.28 &
      0.11 &
      0.17 &
      0.49 &
      0.26 &
      0.27 &
      0.27 &
      58 sec
      \\
\cline{2-9}   &
      Lemma-$\delta$ &
      0.31 &
      0.15 &
      0.14 &
      0.48 &
      0.28 &
      0.27 &
      0.18 &
      9 min 41 sec
      \\
\cline{2-9}     &
      tf-idf $+$ GM &
      0.53 &
      0.37 &
      0.35 &
      0.49 &
      0.42 &
      0.45 &
      0.43 &
      41 min 25 sec
      \\
\cline{2-9}     &
      tf-idf $+$ AC &
      \cellcolor{blue!20}\underline{0.57} &
      \cellcolor{blue!20}\underline{0.42} &
      0.38 &
      0.49 &
      0.45 &
      \cellcolor{blue!20}\underline{0.49} &
      0.46 &
      8 min 5 sec
      \\
\cline{2-9}     &
      s-bert $+$ GM &
      0.43 &
      0.39 &
      \cellcolor{blue!20}\underline{0.40} &
      \cellcolor{blue!20}\underline{0.54} &
      0.43 &
      0.46 &
      0.44 &
      46 min 18 sec
      \\
\cline{2-9}     &
      s-bert $+$ AC &
      0.51 &
      0.42 &
      \cellcolor{blue!20}\underline{0.40} &
      \cellcolor{blue!20}\underline{0.54} &
      \cellcolor{blue!20}\underline{0.46} &
      0.48 &
      \cellcolor{blue!20}\underline{0.47} &
      11 min 15 sec
      \\
\cline{2-10}     &
      \multicolumn{9}{|c|}{$+$ tag embedding}
      \\
\cline{2-10}     &
      tf-idf $+$ GM &
      0.74 &
      0.52 &
      0.40 &
      0.63 &
      0.56 &
      0.59 &
      0.57 &
      43 min 23 sec
      \\
\cline{2-9}     &
      tf-idf $+$ AC &
      0.72 &
      0.51 &
      {0.48} &
      0.64 &
      0.57 &
      0.61 &
      0.59 &
      9 min 27 sec
      \\
\cline{2-9}     &
      S-bert $+$ GM &
      0.74 &
      0.41 &
      0.34 &
      0.67 &
      0.51 &
      0.57 &
      0.54 &
      47 min 12 sec
      \\
\cline{2-9}     &
      s-bert $+$ AC &
      {0.82} &
      \cellcolor{green!20}\textbf{0.53} &
      0.44 &
      \cellcolor{green!20}\textbf{0.72} &
      {0.60} &
      \cellcolor{green!20}\textbf{0.66} &
      {0.63} &
      11 min 42 sec
      \\
\cline{2-9}     &
      m-pair model &
      \cellcolor{green!20}\textbf{0.96} &
      0.42 &
      \cellcolor{green!20}\textbf{0.78} &
      0.35 &
      \cellcolor{green!20}\textbf{0.82} &
      0.65 &
      \cellcolor{green!20}\textbf{0.64} &
      2 hr 11 min 40 sec
      \\
      \hline
      \hline
    \multicolumn{1}{|c|}{\multirow{12}{*}{\rotatebox[origin=c]{90}{\textbf{COVID-19}}}} &
      Lemma &
      0.55 &
      0.39 &
      0.28 &
      0.55 &
      0.51 &
      0.42 &
      0.44 &
      9 sec
      \\
\cline{2-9}   &
      Lemma-$\delta$ &
      0.34 &
      0.29 &
      0.25 &
      0.51 &
      0.35 &
      0.34 &
      0.35 &
      1 min 8 sec
      \\
\cline{2-9}     &
      tf-idf $+$ GM &
      {0.56} &
      {0.41} &
      \cellcolor{blue!20}\underline{0.36} &
      {0.60} &
      0.47 &
      0.50 &
      0.48 &
      6 min 37 sec
      \\
\cline{2-9}     &
      tf-idf $+$ AC &
      0.59 &
      \cellcolor{blue!20}\underline{0.45} &
      \cellcolor{blue!20}\underline{0.36} &
      \cellcolor{blue!20}\underline{0.62} &
      \cellcolor{blue!20}\underline{0.49} &
      \cellcolor{blue!20}\underline{0.54} &
      \cellcolor{blue!20}\underline{0.51} &
      1 min 44 sec 
      \\
\cline{2-9}     &
      s-bert $+$ GM &
      \cellcolor{blue!20}\underline{0.63} &
      \cellcolor{blue!20}\underline{0.45} &
      {0.32} &
      {0.57} &
      0.47 &
      0.51 &
      0.49 &
      8 min 41 sec
      \\
\cline{2-9}     &
      s-bert $+$ AC &
      {0.61} &
      {0.44} &
      {0.35} &
     {0.57} &
      0.48 &
      0.50 &
      0.49 &
      2 min 25 sec
      \\
\cline{2-10}     &
      \multicolumn{9}{|c|}{$+$ tag embedding}
      \\
\cline{2-10}     &
      tf-idf $+$ GM &
      {0.44} &
      {0.33} &
      {0.28} &
      {0.54} &
      0.39 &
      0.40 &
      0.39 &
      7 min 31 sec
      \\
\cline{2-9}     &
      tf-idf $+$ AC &
      0.44 &
      0.34 &
      0.32 &
      0.44 &
      0.4 &
      0.42 &
      0.41 &
      2 min 38 sec
      \\
\cline{2-9}     &
      s-bert $+$ GM &
      0.57 &
      0.41 &
      0.35 &
      0.59 &
      0.47 &
      0.49 &
      0.48 &
      9 min 35 sec
      \\
\cline{2-9}     &
      s-bert $+$ AC &
      {0.63} &
      {0.46} &
      {0.39} &
      {0.59} &
      {0.51} &
      {0.52} &
      {0.52} &
      3 min 19 sec 
      \\
\cline{2-9}     &
      m-pair model &
      \cellcolor{green!20}\textbf{0.86} &
      \cellcolor{green!20}\textbf{0.80} &
      \cellcolor{green!20}\textbf{0.97} &
      \cellcolor{green!20}\textbf{0.65} &
      \cellcolor{green!20}\textbf{0.80} &
      \cellcolor{green!20}\textbf{0.84} &
      \cellcolor{green!20}\textbf{0.82} &
      29 min 18 sec
      \\
      \hline
    \end{tabular}%
\caption{\footnotesize Sentence coreference results before and after tag embedding. GM: Gaussian Mixture based clustering; AC: Agglomerative Clustering; s-bert: sentence-bert; m-pair: supervised mention-pair model. Best results including the tag embedding are marked in boldface and highlighted in green cells. Best results excluding the tag embedding are marked by underline and highlighted in blue cells.}
  \label{tab:coref_resolution}%
\end{table*}%

\noindent\textit{\textbf{Evaluation of coreference resolution}}: For the evaluation of coreference resolution we use several coreference resolution metrics to analyse the model performance. It is apparent from Table \ref{tab:coref_resolution} that the approach based on clustering with \textit{sentence-bert} embeddings by far outperforms the baselines \textit{lemma} and \textit{lemma}-$\delta$. For the CWMG dataset, \textit{sentence-bert} + agglomerative clustering is the best overall; for the other two datasets no single method is a clear winner. However, the primary point that we wish to emphasize in the table is the result after incorporating tag embedding. It can be clearly observed that this intuitive, albeit hitherto unreported, technique almost always produces better results (see Appendix~\ref{tag_creation} and the  Table~\ref{tab:example_tag} therein describing the tag generation process in more details). In fact, the assimilation of the tag embeddings with the \emph{sentence-bert} embeddings boosted the overall F1-score by 13\%, and 16\% for the CWMG and the CWAL datasets respectively. \textcolor{black}{Note that these results hold even if the manual filtering step in the tag generation is completely omitted (see Table~\ref{tab:coref_resolution_unfiltered}}). An interesting observation is that the benefit of the tag embedding is best leveraged by the sentence-bert + agglomerative clustering. For the COVID-19 dataset, since search results are generic, the benefit of tag embedding is less. Furthermore, the supervised model consistently outperforms the unsupervised results across all three datasets. \textcolor{black}{Note that the tag generation is done only once and therefore takes a fixed amount of time. It took 3.26 seconds, 3.47 seconds, and 1.96 seconds per sentence on average to generate knowledge-based tags for CWMG, CWAL, and COVID-19 datasets respectively. The time that the model takes to inference in presence of the tag embeddings is negligible as compared to the model without these embeddings (see the last column of Table \ref{tab:coref_resolution}). For the supervised models though, the major chunk of time is required for the mention pair generation.}\\
\noindent \textit{\textbf{Full system evaluation}}: So far, the assessment for the two components was carried out separately, i.e., the evaluation for the important sentence extraction was based on Level I annotated data while the evaluation for sentence coreference resolution was on the basis of Level II annotations independently. We also conduct the full system evaluation for CWMG and CWAL datasets, i.e., the complete evaluation was only dependent on Level II annotated data. For this case we trained the GAN-BERT classifier with 30\% of the labeled data along with the unlabeled data (discussed in section \ref{Data_annotation}), and had predictions for the rest of 70\% data. Now, we consider only the \textit{true positives} (labeled as important, and also predicted important), before performing the coreference resolution. This task is evaluated based on the Level II annotated data. 
\textcolor{black}{The primary reasons for considering only true positive samples are - (1) we do not have ground-truth Level II annotated data for the non-important sentences (i.e., the false positives), (2) for all practical purposes we are only interested in the coreferences present in the positive predictions (i.e., in the predicted important sentences).} Table \ref{tab:full_system_eval} shows the comparison between the full system evaluation result and the standard result (see Appendix~\ref{appendix:full_system_eval}  for results w/o tags). The results shown here are the average value of the four different standard metrics (MUC, B\textsuperscript{3}, CEAF\_E and BLANC) corresponding to the best performing unsupervised model \textcolor{black}{as well as the mention-pair based supervised model.}\\
\noindent\textit{\textbf{Comparison with TLS}}: \textcolor{black}{Since our method has some parallels with TLS, in this section we perform a thorough comparison with state-of-the-art TLS systems. Note that the output of our system is not similar to that of the standard TLS output. In order to make the comparison possible and fair we added a simple summarization step at the end of our pipeline. We used the BERT extractive summarizer~\cite{miller2019leveraging} to extract the two most important sentences as the summary for each of the fact clusters generated by our method. We evaluated the summaries using the alignment-based ROUGE (AR) F-Score~\cite{martschat-markert-2017-improving}. Unlike~\cite{gholipour-ghalandari-ifrim-2020-examining}, we did not use any date ranking method to rank the dates of the predicted timeline and compared the ground-truth with the top-$k$ predicted timeline. We tested all the approaches using our Level I annotated data as the ground-truth reference. Table \ref{comparison_TLS} shows the detailed comparison of our approach with few of the existing state-of-the-art TLS approaches on two of our datasets. In order to perform these experiments we considered pre-selected 41 formal letters from CWMG in the time period 1930-1935 with more than 1000 words and all the documents of volume 2 from CWAL (from which the Level I annotations were performed) and directly passed through the TLS pipeline using the codes provided by the respective authors. In order to make the comparison further fair, we also performed an experiment by first carrying out important sentence classification using our method and then feeding the filtered data into the TLS pipeline provided by the authors. In order to benefit the TLS models the fact detection for this pre-filtering was performed using the model fine-tuned on our dataset. This modification results in superior performance of the TLS. 
    In fact, fact detection prior to summarization always helps -- our method as well as one of the baseline methods~\cite{gholipour-ghalandari-ifrim-2020-examining} where fact detection can be easily incorporated show significantly\footnote{Statistical significance were performed using Mann–Whitney U test \cite{10.1214/aoms/1177730491}} improved performance. In Table~\ref{full_comparison_TLS} of Appendix~\ref{full_comp_tls} we also show that this fact detection step brings benefits to a standard TLS dataset which has not been built from historical text. The reason for this inferior performance could be that the summary in the standard TLS approaches are highly sensitive to the keywords used for the particular dataset and generating quality keywords for a dataset consisting of diverse facts like ours requires domain-expertise (see Table~\ref{sample_comparison_TLS} in Appendix~\ref{example_summary}).}

\begin{table}[htbp]
\begin{minipage}{0.35\textwidth}
  \centering
  \renewcommand{\arraystretch}{0.8}
  \scriptsize
    \begin{tabular}{|c|c|c|c|c|c|}
    \hline
    Dataset & Type & M & R & P & F1 \\
    \hline
    \multirow{4}{*}{CWMG} & \multirow{2}{*}{Su} & MA & 0.83 & 0.69 & 0.72 \\
\cline{3-6}      &   & MP & 0.74 & 0.63 & 0.64 \\
\cline{2-6}      & \multirow{2}{*}{Un} & MA & 0.65 & 0.71 & 0.68 \\
\cline{3-6}      &   & MP & 0.62 & 0.65 & 0.63 \\
    \hline
    \multirow{4}{*}{CWAL} & \multirow{2}{*}{Su} & MA &  0.82 & 0.65  & 0.64 \\
\cline{3-6}      &   & MP & 0.74  & 0.59  & 0.60\\
\cline{2-6}      & \multirow{2}{*}{Un} & MA & 0.60 & 0.66 & 0.63 \\
\cline{3-6}      &   & MP & 0.55 & 0.59 & 0.57 \\
    \hline
    \end{tabular}%
    \caption{\footnotesize Full system evaluation result. Type: Coref-resolution type, MA: Important sentences obtained through manual annotation, MP: Important sentences obtained from model prediction, Su: Supervised, Un: Unsupervised. Appendix \ref{appendix:full_system_eval} shows the same results without using tag embeddings.}
  \label{tab:full_system_eval}%
\end{minipage}
\begin{minipage}{0.65\textwidth}
\centering
  \renewcommand{\arraystretch}{0.9}
  \scriptsize
    \begin{tabular}{|m{1.5cm}|c|c|c|c|}
    \hline
    \multirow{2}{*}{System} &  \multicolumn{2}{c|}{CWMG Dataset} & \multicolumn{2}{c|}{CWAL Dataset}\\
\cline{2-5}      & AR1-F & AR2-F & AR1-F & AR2-F\\
    \hline
    MM & 0.023 & 0.001 & 0.052 & 0.024\\
    \hline
    DT & 0.008 & 0.001 & 0.022 & 0.002\\
    \hline
    FD (our) + DT   & 0.015* & 0.006* & 0.026* & 0.002\\
    
    \hline
    CLUST & 0.028 & 0.02 & 0.055 & 0.040\\
    
    \hline
    FD (our) + CLUST  & 0.034• & 0.025• & 0.086• & 0.071•\\

    \hline
    Our method & \cellcolor{green!20}\textbf{0.062†*•} & \cellcolor{green!20}\textbf{0.043†*•} & \cellcolor{green!20}\textbf{0.069†*•} & \cellcolor{green!20}\textbf{0.042†*•}\\
    \hline
    \end{tabular}%
\caption{\footnotesize Comparison of our method for the with the existing state-of-the-art TLS methods - (1) MM (submodularity based method): \cite{martschat-markert-2018-temporally} and (2) DT: datewise and (3) CLUST: clustering based TLS by \cite{gholipour-ghalandari-ifrim-2020-examining}, FD: Fact detection. †, *, • show that our results are significantly different from MM, FD + DT, FD + CLUST respectively. In turn, any method with FD (*, •) is significantly better than MM.}
  \label{comparison_TLS}
\end{minipage}
\end{table}%


\if{0}
\begin{table*}[htbp]
  \centering
  \renewcommand{\arraystretch}{0.8}
  \scriptsize
    \begin{tabular}{|c|c|c|c|c|c|c|c|c|c|c|}
    \hline
    \multirow{2}{*}{System} & \multicolumn{2}{c|}{T17 Dataset} & \multicolumn{2}{c|}{Crisis Dataset} & \multicolumn{2}{c|}{Entities Dataset} & \multicolumn{2}{c|}{CWMG Dataset} & \multicolumn{2}{c|}{CWAL Dataset}\\
\cline{2-11}      & AR1-F & AR2-F & AR1-F & AR2-F & AR1-F & AR2-F & AR1-F & AR2-F & AR1-F & AR2-F\\
    \hline
    \textbf{\cite{martschat-markert-2018-temporally}}  & 0.105 & 0.03 & 0.075 & 0.016 & 0.042 & 0.009 & 0.023 & 0.001 & 0.052 & 0.024\\
    \hline
    \textbf{Datewise}  & 0.12 & 0.035 & 0.089 & 0.026 & 0.057 & 0.017 & 0.008 & 0.001 & 0.022 & 0.002\\
    \hline
    \textbf{fact-detection + Datewise}  & 0.122 & 0.039* & 0.092* & 0.03* & 0.042 & 0.017 & 0.015* & 0.006* & 0.026* & 0.002\\
    
    \hline
    \textbf{CLUST}  & 0.082 & 0.02 & 0.061 & 0.013 & 0.051 & 0.015 & 0.028 & 0.02 & 0.055 & 0.04\\
    
    \hline
    \textbf{fact-detection + CLUST}  & 0.085• & 0.026• & 0.062 & 0.012 & 0.054• & 0.017• & 0.034• & 0.025• & 0.086• & 0.071•\\

    \hline
    \textbf{Two step approach (Our method)} & 0.066 & 0.025 & 0.033 & 0.008 & 0.021 & 0.011 & \textbf{0.062†*•} & \textbf{0.043†*•} & \textbf{0.069†*•} & \textbf{0.042†*•}\\
    \hline
    \end{tabular}%
\caption{\footnotesize Comparison of our method for the TLS task on the 3 standard news TLS datasets as well as on our dataset, with the existing state of the art TLS methods - (1) submodularity based method by \cite{martschat-markert-2018-temporally} and (2) datewise and (3) CLUST: clustering based TLS by \cite{gholipour-ghalandari-ifrim-2020-examining}. † denotes significant\footnote{Statistical significance were performed using Mann–Whitney U test \cite{10.1214/aoms/1177730491}} improvement over \cite{martschat-markert-2018-temporally}, * over DT, and  • over CLUST}
  \label{comparison_TLS}%
\end{table*}%

\begin{table}[t]
  \centering
  \renewcommand{\arraystretch}{0.9}
  \scriptsize
    \begin{tabular}{c|c|c|c|c}
    \hline
    \multirow{2}{*}{System} &  \multicolumn{2}{c|}{CWMG Dataset} & \multicolumn{2}{c}{CWAL Dataset}\\
\cline{2-5}      & AR1-F & AR2-F & AR1-F & AR2-F\\
    \hline
    MM & 0.023 & 0.001 & 0.052 & 0.024\\
    \hline
    DT & 0.008 & 0.001 & 0.022 & 0.002\\
    \hline
    FD (our) + DT   & 0.015* & 0.006* & 0.026* & 0.002\\
    
    \hline
    CLUST & 0.028 & 0.02 & 0.055 & 0.040\\
    
    \hline
    FD (our) + CLUST  & 0.034• & 0.025• & 0.086• & 0.071•\\

    \hline
    Our method & \cellcolor{green!20}\textbf{0.062†*•} & \cellcolor{green!20}\textbf{0.043†*•} & \cellcolor{green!20}\textbf{0.069†*•} & \cellcolor{green!20}\textbf{0.042†*•}\\
    \hline
    \end{tabular}%
\caption{\footnotesize Comparison of our method for the with the existing state-of-the-art TLS methods - (1) MM (submodularity based method): \cite{martschat-markert-2018-temporally} and (2) DT: datewise and (3) CLUST: clustering based TLS by \cite{gholipour-ghalandari-ifrim-2020-examining}, FD: Fact detection. †, *, • show that our results are significantly different from MM, FD + DT, FD + CLUST respectively. In turn, any method with FD (*, •) is significantly better than MM.}
  \label{comparison_TLS}%
\end{table}%
\fi

\section{Ablation study}
\textcolor{black}{We performed two ablation studies - first, to check the effectiveness of manual filtering of noisy tags, second, to assess the added value of each component in the mention-pair model.}

\noindent\textbf{Sentence coreference resolution results without manual filtering of tags}:
\label{event_coref_resolution_without_manual_filetring}
\textcolor{black}{Table \ref{tab:coref_resolution_unfiltered} shows result obtained from different coreference resolution techniques when we do not include any manual filtering steps to the generated tags. It can be noticed that there is not much difference in the results even when we omit this step.} 

\noindent\textbf{Added value of each element in the mention-pair model}:
Table~\ref{tab:Added_value} shows the added value of each feature in the mention-pair model. For both the historical texts we observe that inclusion of each feature improves the overall performance. The best improvement is observed on the inclusion of the external knowledge in the form of tag embeddings.

\begin{table}[htbp]
\begin{minipage}{0.55\textwidth}
\renewcommand{\arraystretch}{1.1}
  \centering
  \scriptsize
    \begin{tabular}{|c|c|c|c|c|c|c|c|c|}
    \hline
    \multicolumn{1}{|c|}{\multirow{2}{*}{D}} &
      \multirow{2}{*}{M} &
      \multicolumn{1}{|c|}{MUC} &
      \multicolumn{1}{|c|}{B$^{3}$} &
      \multicolumn{1}{|c|}{C} &
      \multicolumn{1}{|c|}{B} &
      \multicolumn{3}{|c|}{Avg (overall)}
      \\
\cline{3-9}     &
      \multicolumn{1}{c|}{} &
      \multicolumn{1}{|c|}{F1} &
      \multicolumn{1}{|c|}{F1} &
      \multicolumn{1}{|c|}{F1} &
      \multicolumn{1}{|c|}{F1} &
      \multicolumn{1}{|c|}{R} &
      \multicolumn{1}{|c|}{P} &
      \multicolumn{1}{|c|}{F1}
      \\
    \hline
    \multicolumn{1}{|c|}{\multirow{5}{*}{\rotatebox[origin=c]{90}{\textbf{CWMG}}}} &

      tf-idf+GM &
      0.61 &
      0.55 &
      0.51 &
      0.58 &
      0.62 &
      0.57&
      0.56
      \\
\cline{2-9}     &
      tf-idf+AC &
      0.64 &
      0.59 &
      0.51 &
      0.66 &
      0.58 &
      0.64 &
      0.60
      \\
\cline{2-9}     &
      s-bert+GM &
      0.68 &
      0.61 &
      0.44 &
      0.63 &
      0.62 &
      0.60 &
      0.59
      \\
\cline{2-9}     &
      s-bert+AC &
      {0.76} &
      {0.71} &
      {0.50} &
      {0.72} &
      {0.65} &
      {0.72} &
      {0.67}
      \\
\cline{2-9}     &
      m-pair &
      {0.92} &
      {0.61} &
      {0.85} &
      {0.53} &
      {0.85} &
      {0.70} &
      {0.73}
      \\
    \hline
    \hline
    \multicolumn{1}{|c|}{\multirow{5}{*}{\rotatebox[origin=c]{90}{\textbf{CWAL}}}} &
    
      tf-idf+GM &
      0.76 &
      0.51 &
      0.44 &
      0.65 &
      0.55 &
      0.59 &
      0.59
      \\
\cline{2-9}     &
      tf-idf + AC &
      0.75 &
      0.50 &
      {0.49} &
      0.65 &
      0.56 &
      0.63 &
      0.59
      \\
\cline{2-9}     &
      S-bert+GM &
      0.76 &
      0.40 &
      0.35 &
      0.69 &
      0.51 &
      0.59 &
      0.55
      \\
\cline{2-9}     &
      s-bert+AC &
      {0.81} &
      {0.59} &
      0.47 &
      {0.70} &
      {0.63} &
      {0.72} &
      {0.64}
      \\
\cline{2-9}     &
      m-pair &
      {0.95} &
      0.43 &
      {0.76} &
      0.36 &
      {0.81} &
      0.67 &
      {0.62}
      \\
      \hline
      \hline
    \multicolumn{1}{|c|}{\multirow{5}{*}{\rotatebox[origin=c]{90}{\textbf{COVID-19}}}} &
      
      tf-idf+GM &
      {0.40} &
      {0.33} &
      {0.26} &
      {0.55} &
      0.39 &
      0.44 &
      0.38
      \\
\cline{2-9}     &
      tf-idf+AC &
      0.42 &
      0.35 &
      0.34 &
      0.43 &
      0.41 &
      0.39 &
      0.38
      \\
\cline{2-9}     &
      s-bert+GM &
      0.56 &
      0.43 &
      0.36 &
      0.57 &
      0.44 &
      0.49 &
      0.48
      \\
\cline{2-9}     &
      s-bert+AC &
      {0.65} &
      {0.44} &
      {0.37} &
      {0.59} &
      {0.52} &
      {0.50} &
      {0.51}
      \\
\cline{2-9}     &
      m-pair &
      {0.84} &
      {0.80} &
      {0.95} &
      {0.66} &
      {0.79} &
      {0.82} &
      {0.81}
      \\
      \hline
    \end{tabular}%
\caption{\footnotesize Sentence coreference results without using manual filtering for the tags. D: dataset, M: model, GM: Gaussian Mixture based clustering; AC: Agglomerative Clustering; s-bert: sentence-bert, m-pair: mention-pair model, B: BLANC, C: CEAF\_E. The results mostly remain unaffected.}
  \label{tab:coref_resolution_unfiltered}%
 \end{minipage}
 \begin{minipage}{0.45\textwidth}
 \centering
 \renewcommand{\arraystretch}{1.1}
 \scriptsize
\begin{tabular}{l|r|r|r}\hline
D &F &Avg F1 &Inc. \\\hline
\multirow{4}{*}{\rotatebox[origin=c]{90}{CWMG}} &S &0.613 &- \\\cline{2-4}
&S+D &0.657 &0.044 \\\cline{2-4}
&S+D+A &0.688 &0.031 \\\cline{2-4}
&S+D+A+T &0.720 &0.038 \\\cline{1-4}
\multirow{4}{*}{\rotatebox[origin=c]{90}{CWAL}} &S &0.394 &- \\\cline{2-4}
&S+D &0.544 &0.15 \\\cline{2-4}
&S+D+A &0.560 &0.016 \\\cline{2-4}
&S+D+A+T &0.640 &0.008 \\\cline{1-4}
\multirow{4}{*}{\rotatebox[origin=c]{90}{Covid-19}} &S &0.791 &- \\\cline{2-4}
&S+D &0.778 &-0.013 \\\cline{2-4}
&S+D+A &0.811 &0.033 \\\cline{2-4}
&S+D+A+T &0.820 &0.009 \\
\hline
\end{tabular}
\caption{Added value of each component in the mention-pair model for each dataset; F: features, S: considering sentence embedding as the only feature, D: date, A: action, T: tag.}\label{tab:Added_value}
\end{minipage}
  
\end{table}%

\section{Timeline visualization}  
Generating a timeline would not be that impactful unless it is visualized in an interpretable and convenient way. We incorporate an elegant visualization for the generated fact/demand timelines using \emph{vis-timeline} javascript library (Appendix \ref{appendix:timeline} shows an example timeline).  

\noindent\textit{Survey}: In order to understand the effectiveness of the interface we ran an online crowd-sourced survey. Out of 33 participants with different educational backgrounds, overall 93\% agreed that the interface was very useful for summarization of historical timeline of facts. 88\% participants found some information which would have been hard for them to fathom just by reading the CWMG plaintext (more results in Appendix \ref{appendix:survey}). 



\if{0}
\begin{table}\centering

\scriptsize
\begin{tabular}{l|r|r|r}\hline
Dataset &Features &Average F1 &Increment \\\hline
\multirow{4}{*}{CWMG} &S &0.613 &- \\\cline{2-4}
&S+Date &0.657 &0.044 \\\cline{2-4}
&S+Date+Action &0.688 &0.031 \\\cline{2-4}
&S+Date+Action+Tag &0.72 &0.038 \\\cline{1-4}
\multirow{4}{*}{CWAL} &S &0.394 &- \\\cline{2-4}
&S+Date &0.544 &0.15 \\\cline{2-4}
&S+Date+Action &0.56 &0.016 \\\cline{2-4}
&S+Date+Action+Tag &0.64 &0.008 \\\cline{1-4}
\multirow{4}{*}{Covid-19} &S &0.791 &- \\\cline{2-4}
&S+Date &0.778 &-0.013 \\\cline{2-4}
&S+Date+Action &0.811 &0.033 \\\cline{2-4}
&S+Date+Action+Tag &0.82 &0.009 \\
\hline
\end{tabular}
\caption{Added value of each element in the mention-pair model for each dataset; S: considering sentence embedding as the only feature.}\label{tab:Added_value}
\end{table}
\fi

\section{Conclusion}
In this work we presented a framework to generate fact timeline from any timestamped document. The entire pipeline has two parts -- important sentence detection and sentence coreference resolution. We achieve very encouraging results for both these tasks. 
\textcolor{black}{While it is true that our evaluations are based on two historical texts, our methods are generic and can be easily extended to other datasets. The system that we developed is not limited to any actor specific fact (human or location) which, in fact, made the coreference resolution task even more challenging.} We believe that our work will open up new and exciting opportunities in history research and education. 

\bibliography{main}
\bibliographystyle{splncs04}


\appendix

\section{Appendices}
\label{sec:appendix}

\subsection{Example timeline of facts}\label{eg:timeline}
\begin{figure}[H]
    \centering
    \includegraphics[scale=0.3]{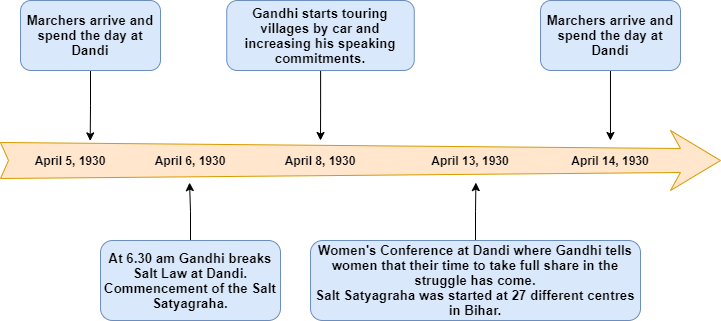}
    \caption{Sample timeline example extracted from documents.}
    \label{event_timeline_intro}
\end{figure}

The method that the we propose can generate a timeline as shown in Figure~\ref{event_timeline_intro}. This can be remarkably helpful to recognize the context and the actors of a particular fact in a certain period.


\subsection{Sample annotations}
\label{appendix:sample_annot}
Table \ref{sample_annot_1} shows the examples of Level I annotated data (sentence classification) and Table \ref{sample_annot_2} illustrates Level II annotated data (coreference resolution) for some portions in the CWMG dataset. 
\begin{table}[H]
    \centering
    \includegraphics[width=0.48\textwidth,height=0.5\textheight,keepaspectratio]{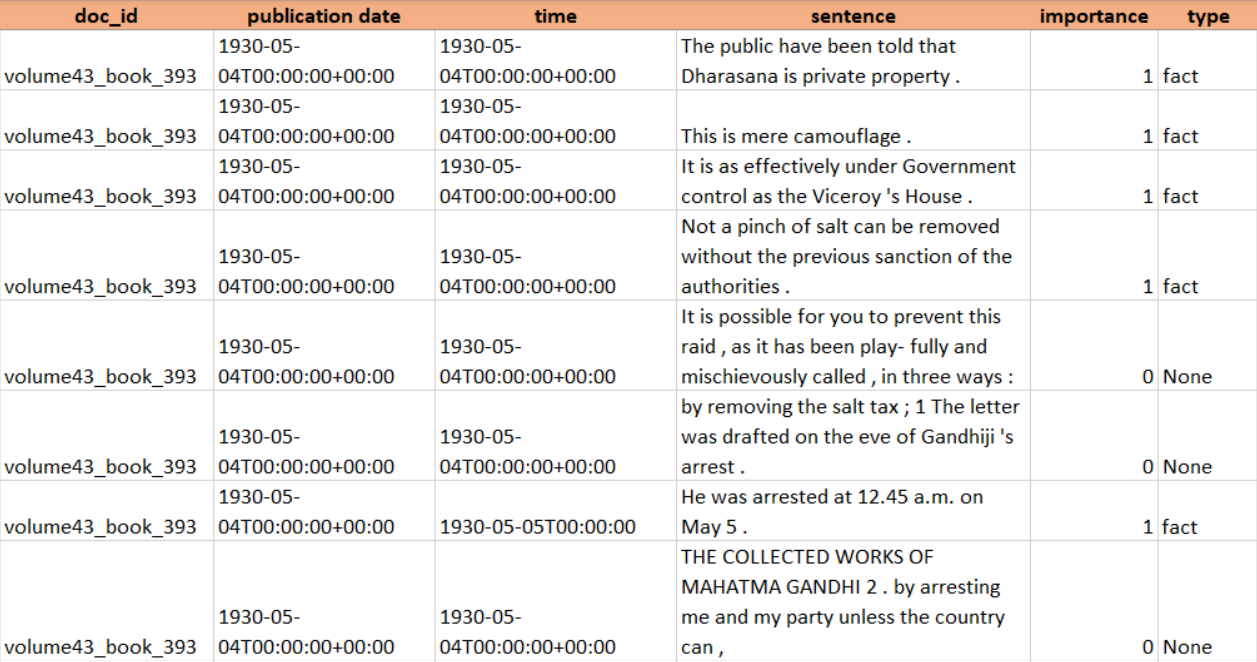}
    \caption{\footnotesize Sample Level I annotation of CWMG dataset (None indicates `other' category).}
    \label{sample_annot_1}
\end{table}

\begin{table}[H]
    \centering
    \includegraphics[width=0.48\textwidth,height=0.5\textheight,keepaspectratio]{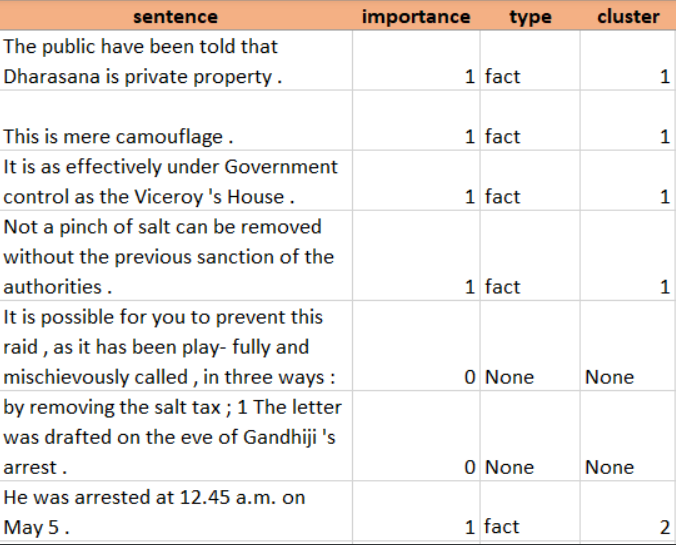}
    \caption{\footnotesize Sample Level II annotation of CWMG dataset. We only marked the cluster value for the sentences which are marked as important by at least 2 annotators during the level I annotation. }
    \label{sample_annot_2}
\end{table}


\subsection{Annotator agreement using different metrics for Level II annotated samples}
\label{appendix:annot_agree}

\begin{table}[H]
\renewcommand{\arraystretch}{0.8}
  \centering
  \footnotesize
    \begin{tabular}{|c|c|c|c|c|}
    \hline
    \multirow{2}{*}{\textbf{Dataset}} & \multicolumn{4}{|c|}{\textbf{Metric}} \\
\cline{2-5}    \multicolumn{1}{|c|}{} & \multicolumn{1}{|c|}{MUC} & \multicolumn{1}{|c|}{B$^{3}$} & \multicolumn{1}{|c|}{CEAF\_E} & \multicolumn{1}{|c|}{BLANC} \\
    \hline
    \textbf{CWMG} & 0.74   & 0.72 & 0.65 & 0.77\\
    \hline
    \textbf{CWAL} & 0.61 & 0.54 & 0.55 & 0.59 \\
    \hline
    \textbf{COVID-19} & 0.78    & 0.81 & 0.71 & 0.74 \\
    \hline
    \end{tabular}%
    \caption{\footnotesize Annotator agreement (F1 score) for Level II annotated data using different metrics.}
  \label{appendix:annot_agreement}%
\end{table}%

\subsection{Details of tag creation method}
\label{tag_creation}
\textcolor{black}{
The generation of tags from world knowledge for a particular sentence is an important part of our pipeline, which contain the manual filtering part. We take the sentence as query, and by using \emph{googlesearch} api we obtain the top 5 retrieved urls and scrape the texts from these. We also consider the original document from where the sentence is being extracted (for COVID-19 data document this is not present) to gather additional context. Based on the internet connectivity, server response time, number of results per page it can take from 1 second to up to a maximum of 30 seconds for scraping the texts from web for each of the sentences. Then we use three methods (\emph{TextRank}, \emph{Rake}, and \emph{pointwise mutual information}) to collect top 5 bigrams (we observed bigrams provide most relevant results) by each of the methods. During the process we automatically filter the stop words, and consider the bigrams which belong to one of the following POS categories - \emph{'JJ', 'JJR', 'JJS', 'NN', 'NNS', 'NNP', 'NNPS'}. The parts of speech tags are determined using \emph{nltk pos\_tag} module. Table \ref{tab:example_tag} shows examples of top 5 tags generated for a sentence by each of the three above methods.} \\
\begin{table*}[ht]
   \centering
  \scriptsize
  
    \begin{tabular}{p{0.35\linewidth}|p{0.15\linewidth}|p{0.35\linewidth}}
\hline    Sentence & \multicolumn{1}{c|}{method} & \multicolumn{1}{c}{example tags} \\
    \hline
    {} & TextRank & government notices', 'government control', 'non violence', 'private salt', 'young india' \\
\cline{2-3} Paddy fields are reported to have been burnt, eatables forcibly taken.     & Rake & without hesitation', 'victims success', 'viceroy house', 'unthinkable cruelties', 'unnecessary bones' \\
\cline{2-3}      & Pointwise Mutual Information & civil disobedience', 'salt tax', 'civil resisters',  'TO VICEROY', 'satyagraha programme' \\
    \hline
    \end{tabular}%
    \caption{Examples of generated tags.}
  \label{tab:example_tag}%
\end{table*}%

\subsection{Architecture diagram of supervised mention-pair model}\label{sup_diag}
\label{appendix:supervised-model-architecture}
\begin{figure}
    \centering
    \includegraphics[width=0.5\textwidth,height=0.5\textheight,keepaspectratio]{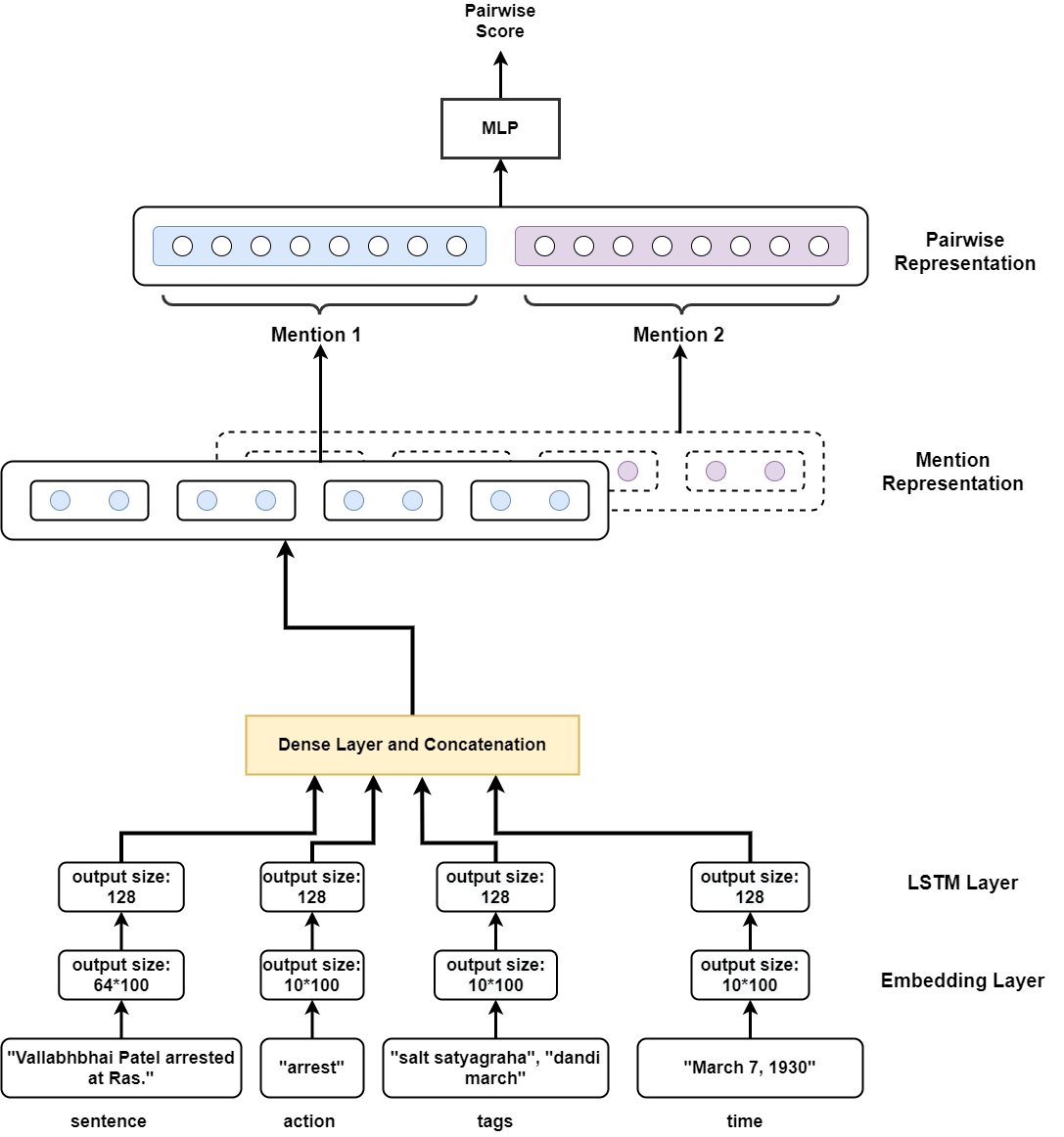}
    \caption{\footnotesize An illustration of the Fact mention-pair model.}
    \label{supervised_architecture}
\end{figure}

 Figure \ref{supervised_architecture} represents the model architecture, which is inspired from \cite{barhom2019revisiting}. 
 
 \subsection{Effectiveness of fact detection in TLS task}\label{full_comp_tls}
 Table \ref{full_comparison_TLS} shows how our fact detection step improves the performance for a standard TLS dataset also which has not been built from historical text. 

\begin{table}[H]
  \centering
  \renewcommand{\arraystretch}{1}
  \scriptsize
    \begin{tabular}{|c|c|c|c|}
    \hline
    \multirow{2}{*}{System} & \multicolumn{2}{c|}{Timeline17 Dataset}\\
\cline{2-3}      & AR1-F & AR2-F \\
    \hline
    MM  & 0.105 & 0.03 \\
    \hline
    DT  & 0.12 & 0.035 \\
    \hline
    FD (our) + DT  & 0.122 & 0.039*  \\
    
    \hline
    CLUST  & 0.082 & 0.02 \\
    
    \hline
    FD (our) + CLUST  & 0.085• & 0.026•  \\

    \hline
    \end{tabular}%
\caption{\footnotesize Comparison of the performance with and without incorporating our fact detection step for the TLS task on a standard TLS dataset. TLS methods used are -- (1) MM (submodularity based method): \cite{martschat-markert-2018-temporally} and (2) DT: datewise and (3) CLUST: clustering based TLS by \cite{gholipour-ghalandari-ifrim-2020-examining}. FD: Our fact detection method. † denotes significant improvement over \cite{martschat-markert-2018-temporally}, * over DT, and  • over CLUST.}
  \label{full_comparison_TLS}%
\end{table}
 
 \subsection{Example summaries}\label{example_summary}
 
 In Table~\ref{sample_comparison_TLS} we present a few examples comparing the summaries produced by our method vis-a-vis the approach outlined in using \cite{gholipour-ghalandari-ifrim-2020-examining}. The blue portions indicate the parts that are present in the ground-truth.
 
 \begin{table}[ht]
     \centering
     \includegraphics[width=0.48\textwidth,height=0.5\textheight,keepaspectratio]{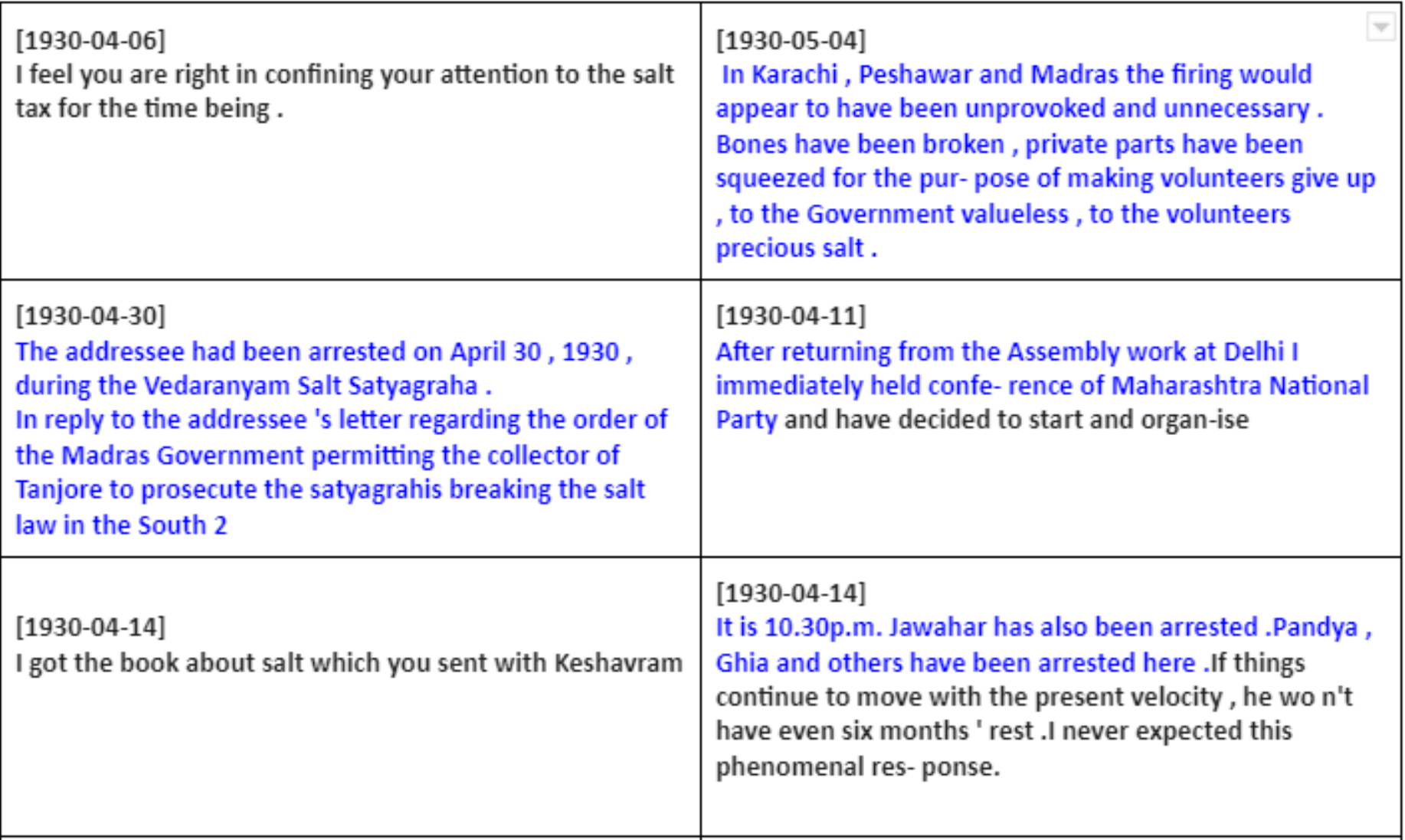}
     \caption{\footnotesize Sample summary generated using \cite{gholipour-ghalandari-ifrim-2020-examining} (left) and our method (right) on the CWMG dataset. Text in \textcolor{black}{blue} indicates the portion present in the ground-truth timeline.}
     \label{sample_comparison_TLS}
 \end{table}

\subsection{Full system evaluation without tags}
\label{appendix:full_system_eval}
Table \ref{tab:full_system_eval_without_tag} shows the coreference resolution results for the full system using both supervised (fact mention-pair model) and unsupervised (s-bert + agglomerative clustering) methods without using external tag embeddings. 

\begin{table}[H]
  \centering
  \renewcommand{\arraystretch}{0.8}
  \scriptsize
    \begin{tabular}{c|c|c|c|c|c}
    \hline
    Dataset & Coref-resolution type & methods & R & P & F1 \\
    \hline
    \multirow{4}{*}{CWMG} & \multirow{2}{*}{Supervised} & MA & 0.76 & 0.65 & 0.68 \\
\cline{3-6}      &   & MP & 0.62 & 0.55 & 0.52 \\
\cline{2-6}      & \multirow{2}{*}{Unsupervised} & MA & 0.55 & 0.56 & 0.55 \\
\cline{3-6}      &   & MP & 0.41 & 0.42 & 0.41 \\
    \hline
    \multirow{4}{*}{CWAL} & \multirow{2}{*}{Supervised} & MA &  0.74 & 0.62  & 0.66 \\
\cline{3-6}      &   & MP & 0.48  & 0.56  & 0.51\\
\cline{2-6}      & \multirow{2}{*}{Unsupervised} & MA & 0.46 & 0.48 & 0.47 \\
\cline{3-6}      &   & MP & 0.31 & 0.30 & 0.31 \\
    \hline
    \end{tabular}%
    \caption{\footnotesize Full system evaluation result without tags. MA: Important sentences obtained through manual annotation, MP: Important sentences obtained from model prediction.}
  \label{tab:full_system_eval_without_tag}%
\end{table}%

\subsection{Sample timeline}
\label{appendix:timeline}
\begin{figure}
    \centering
    \includegraphics[width=0.7\textwidth,height=0.5\textheight,keepaspectratio]{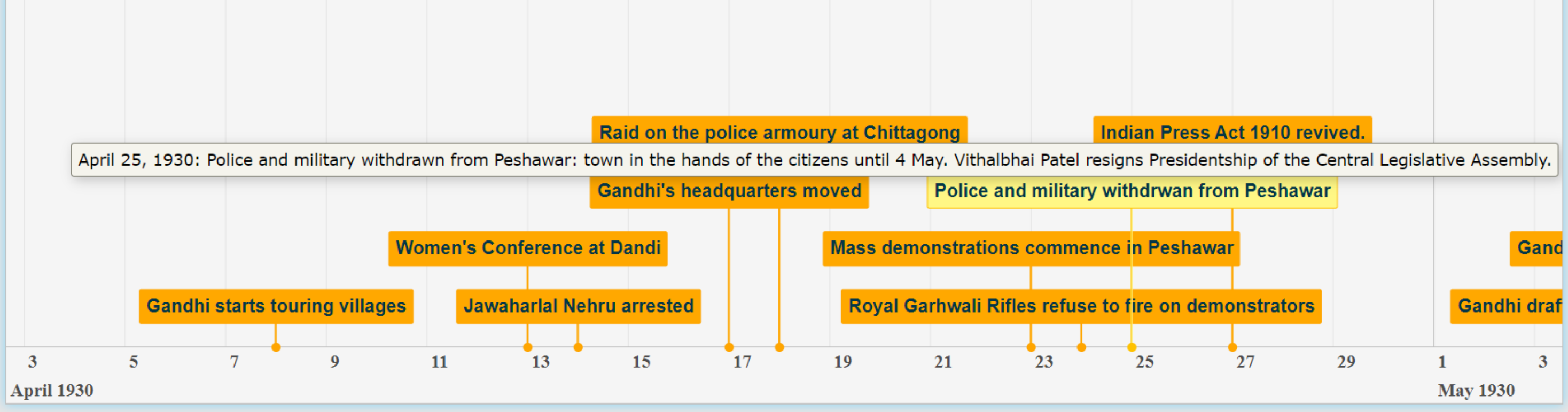}
    \caption{\footnotesize Sample visualization of timeline generated from the CWMG dataset.}
    \label{cwmg_visualization}
\end{figure}

After resolving the sentence coreference, the generated data is used to create the timeline. In order to generate the title for a specific event, we have used BERT extractive summarizer~\cite{miller2019leveraging}. 
The idea of visualisation was to make the tool accessible to historians as well as run a survey of the utility of the tool in the first place. Figure \ref{cwmg_visualization} shows a sample timeline generated by the tool from the CWMG dataset.

\subsection{Online survey}
\label{appendix:survey}

In the survey we asked participants a number of questions regarding the readability, correctness and relevance about the information in the generated timeline. 33 participants with various educational backgrounds took part in the survey. 79\% of the participants noted that the interface was easily readable. 73\% of the total participants reported that they were very satisfied with the overall quality of the automatically generated timeline summaries.

\subsection{Ethical considerations}
\label{appendix:ethical_consideration}
We have framed our datasets by collecting textual information from publicly available online resources and these do not contain any individual private information. The two historical datasets, i.e., the CWMG and the CWAL have been constructed by using the two specific online sources mentioned in \ref{Dataset}, while the privacy rights have been acknowledged. The contents in the COVID-19 event dataset are collected from freely accessible Wikipedia and publicly available information from \url{https://who.int}. Further, the datasets have been annotated by the research scholars and university undergraduate students voluntarily. \textcolor{black}{Finally, in order to avoid concerns of bias in the survey we had 5 expert historians out of the 33 participants. Three among these participants found the information on the timeline fully correct and the other two found it mostly correct. Further four of them agreed that the sentences appeared in the timeline are important for summarizing the life events. Since the observations of the experts align very well with nontechnical audience, we are confident that the accuracy and factuality of the information gathered and shown on the timeline are not misleading.}


%
%
%
%




\end{document}